\documentclass[10pt,twocolumn,letterpaper]{article}

\usepackage{wacv}              
\usepackage{multirow}
\usepackage{float}

\definecolor{wacvblue}{rgb}{0.21,0.49,0.74}
\usepackage[pagebackref,breaklinks,colorlinks,allcolors=wacvblue]{hyperref}

\definecolor{revblue}{rgb}{0.0, 0.0, 0.7}

\definecolor{mypink}{rgb}{0.0, 0.0, 0.0}
\newcommand{\news}[1]{\textcolor{mypink}{#1}}

\def\wacvPaperID{1328} 
\def\confName{WACV}
\def\confYear{2026}

\title{IPTQ-ViT: Post-Training Quantization of Non-linear Functions \\ for Integer-only Vision Transformers}

\author{Gihwan Kim$^{1}$ \quad
        Jemin Lee$^{2}$ \quad
        Hyungshin Kim$^{1}$\\[0.3em]
        $^{1}$Chungnam National University, Daejeon, Republic of Korea\\
        $^{2}$Electronics and Telecommunications Research Institute (ETRI), Daejeon, Republic of Korea\\[0.3em]
        {\tt\small gihwan.kim98@o.cnu.ac.kr \quad leejaymin@etri.re.kr \quad hyungshin@cnu.ac.kr}
}

\begin{document}
\maketitle
\begin{abstract}
Previous Quantization-Aware Training (QAT) methods for vision transformers rely on expensive retraining to recover accuracy loss in non-linear layer quantization, limiting their use in resource-constrained environments. In contrast, existing Post-Training Quantization (PTQ) methods either partially quantize non-linear functions or adjust activation distributions to maintain accuracy but fail to achieve fully integer-only inference. 
In this paper, we introduce IPTQ-ViT, a novel PTQ framework for fully integer-only vision transformers without retraining. We present approximation functions: a polynomial-based GELU optimized for vision data and a bit-shifting-based Softmax designed to improve approximation accuracy in PTQ. In addition, we propose a unified metric integrating quantization sensitivity, perturbation, and computational cost to select the optimal approximation function per activation layer.
IPTQ-ViT outperforms previous PTQ methods, achieving up to 6.44\%p (avg. 1.78\%p) top-1 accuracy improvement for image classification, 1.0 mAP for object detection. 
IPTQ-ViT outperforms partial floating-point PTQ methods under W8A8 and W4A8, and achieves accuracy and latency comparable to integer-only QAT methods. We plan to release our code\footnote{\url{https://github.com/gihwan-kim/IPTQ-ViT.git}}.
\end{abstract}

\section{Introduction}
\label{sec:intro}
Vision Transformers (ViTs)~\cite{vit} have achieved state-of-the-art performance in various computer vision tasks~\cite{detr, chen2021crossvit, swin, segmenter, vit-yolo, deformable-detr, zhou2023octr}. However, their large model size and computational complexity pose considerable challenges for deployment on resource-constrained devices and mobile environments~\cite{survey_1}. To address these challenges, model compression and inference acceleration techniques have been actively studied. Among these techniques, quantization methods that lower parameter precision effectively reduce both model size and computational cost~\cite{survey_1, intro-model-quantization-01, survey_2}.

Integer-only quantization improves computational efficiency by replacing floating-point operations with integer arithmetic, thereby reducing data transfer, eliminating dequantization overhead, and fully utilizing hardware-efficient integer units~\cite{why-integer, hawq-v3-cnn, i-vit, i-bert}. While this approach has shown strong success in Convolutional Neural Networks (CNNs) that primarily consist of linear operations~\cite{uniform-quantization-jacob, hawq-v3-cnn}, applying it to vision transformers remains challenging. Vision transformers rely on non-linear functions such as Softmax, LayerNorm, and GELU, which are not naturally compatible with integer arithmetic and often require approximation or fake quantization for deployment. In addition, activation layers in vision transformers typically exhibit long-tailed and imbalanced distributions, making them highly sensitive to quantization and leading to performance degradation~\cite{repq-vit, fq-vit, ptq4vit, noisyquant}.

To mitigate these issues, QAT methods~\cite{mixed_v1, i-vit, i-bert} approximate all non-linear functions with integer-friendly operations and reduce quantization errors through retraining. However, QAT methods rely on high-performance GPUs and incur significant retraining overhead which limit their applicability in real-world deployments.
In contrast, PTQ methods~\cite{mixed_v1, psaq-vit-v1, psaq-vit-v2, fq-vit, repq-vit, noisyquant, clamp-vit, ptq4vit} are deployment friendly as they enable the quantization of pretrained models without retraining. PTQ methods, however, struggle to mitigate the accuracy degradation caused by quantizing non-linear operations without retraining.
Furthermore, existing PTQ approaches typically rely on partial or fake quantization for activation layers, failing to achieve fully integer-only inference. 
One notable attempt is FQ-ViT~\cite{fq-vit}, a state-of-the-art PTQ method that applies quantization to LayerNorm and Softmax to enable integer-only inference. However, since GELU remains in floating point, it does not achieve a fully integer-only vision transformer.

In this paper, we present IPTQ-ViT, a novel PTQ framework for fully integer-only vision transformers that effectively mitigates accuracy degradation in PTQ caused by quantized non-linear operations without retraining. 
QAT-based approximation functions, such as I-ViT~\cite{i-vit} and I-BERT~\cite{i-bert}, can be directly applied to PTQ for vision transformers to address non-linear quantization challenges.
However, we observe that heuristically applying these functions in PTQ leads to severe accuracy degradation, with accuracy dropping to as low as 0.08\% in extreme cases (see Tab.~\ref{tab:motivation_qat2ptq_acc}). 
We analyze the reasons behind this accuracy drop and propose new approximation functions tailored to PTQ. 
Kim et al.~\cite{mixed_v1} select approximation functions based on quantization sensitivity in QAT.
We extend this idea to PTQ without retraining, thereby reducing analysis overhead.
Our method also expands the function-assignment search space to enhance flexibility and accuracy. To support this, we propose a unified metric that integrates quantization sensitivity, perturbation, and computational cost, enabling efficient layer-wise function assignment.

Our main contributions are as follows:
\begin{enumerate}[itemsep=0pt, topsep=0pt]
    \item We introduce a novel PTQ framework for integer-only vision transformers, fully quantizing both linear and non-linear operations without retraining. 
    \item We propose new approximation functions tailored for PTQ: \textit{Data-aware Poly-GELU} optimized for vision data and \textit{Efficient Bit-Softmax} which improves approximation accuracy.
    \item We design \textit{Unified Metric} that integrates quantization sensitivity, perturbation, and computational complexity to assign optimal approximation functions per activation layer.
    \item We conduct extensive experiments on image classification, object detection, and latency evaluation. IPTQ-ViT outperforms existing PTQ methods and achieves accuracy and latency comparable to QAT methods. 
\end{enumerate}
\section{Related Work}
\label{sec:related_work}

\textbf{PTQ for Vision Transformers. } 
PTQ methods have been proposed to efficiently deploy Vision Transformers on resource-constrained devices~\cite{survey_1}. 
Previous studies focus on correcting activation distribution distortions that negatively impact quantization performance~\cite{repq-vit, fq-vit, ptq4vit, zhong2024erq}. RepQ-ViT~\cite{repq-vit} and FQ-ViT~\cite{fq-vit} tackle channel-wise variance in LayerNorm and imbalance in attention maps. PTQ4ViT~\cite{ptq4vit} separates GELU and Softmax using Twin-Uniform quantization, while ERQ~\cite{zhong2024erq} corrects quantization errors via ridge regression. Recent works, data-free PTQ~\cite{psaq-vit-v1, psaq-vit-v2, clamp-vit}, explore synthetic calibration data to eliminate reliance on real data. 
However, these PTQ methods rely on partial floating-point operations or fail to fully quantize activation layers, limiting fully integer-only inference.

\par\noindent
\textbf{Non-linear Operation Quantization.} 
Integer-only quantization replaces floating-point operations with integer arithmetic, offering significant efficiency gains~\cite{why-integer}, and has been successfully applied to CNNs~\cite{uniform-quantization-jacob, why-integer}.
However, applying integer-only quantization to ViTs and Large Language Models (LLMs) presents challenges due to the non-linearity of activation layers such as Softmax, GELU, and LayerNorm.
I-BERT~\cite{i-bert} proposes polynomial approximations for GELU and Softmax in language models. FQ-ViT~\cite{fq-vit} proposes log2 quantization and i-exp~\cite{i-bert} to approximate Softmax, utilizing power-of-two scaling for LayerNorm in PTQ. ShiftAddLLM~\cite{you2024shiftaddllm} streamlines these approximations to utilize only shift-add operations. I-ViT~\cite{i-vit} approximates Softmax and GELU using bit-shifting and corrects quantization error with retraining, while Kim et al.~\cite{mixed_v1} select optimal functions per layer using SQNR-based metrics in QAT. I-LLM~\cite{illm} extends bit-shifting approximation to LLMs. 

\par\noindent
\textbf{Limitations of Non-linear Function Quantization. } 
The methods~\cite{illm, i-bert, qin2025picachu, you2024shiftaddllm} tailored for LLMs often fail to generalize to vision tasks, as evidenced by the performance degradation of I-BERT~\cite{i-bert} when applied to QAT-based ViTs~\cite{i-bert}.
QAT-based ViT methods~\cite{mixed_v1, i-bert, i-vit} require significant retraining costs, powerful GPU resources, and the accessibility of the complete training dataset, which limits their applicability in deployment.
PTQ-based methods~\cite{fq-vit} offer greater ease of deployment; however, they do not involve retraining, resulting in increased sensitivity to approximation error. 

\begin{figure*}[t]
\centering
\includegraphics[scale=0.26]{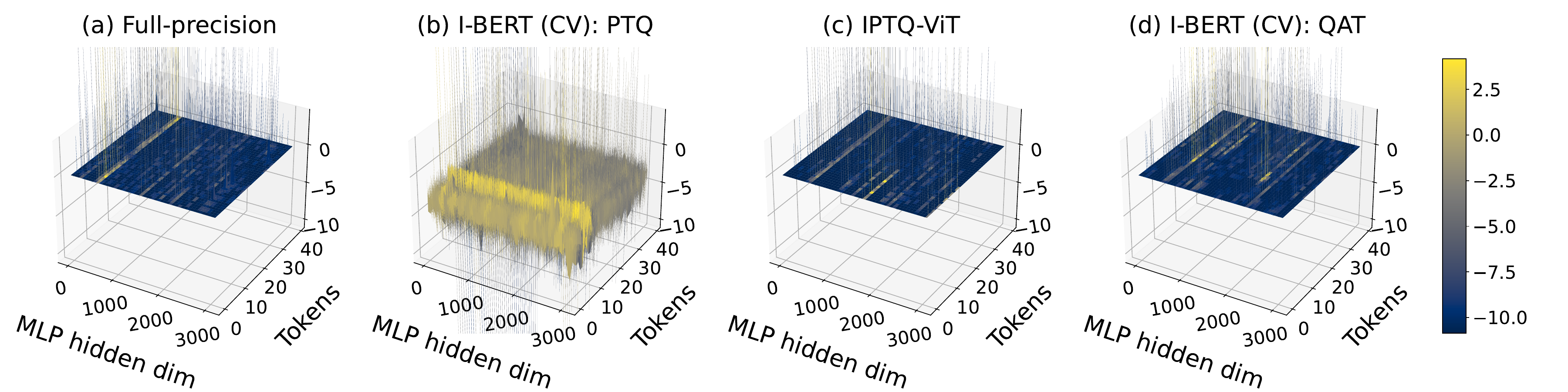}
\caption{
Activation distributions of GELU in the 11-th block of \emph{ViT-B}, which shows the highest quantization sensitivity in Tab.~\ref{tab:motivation_qat2ptq_sensitivity} (I-BERT${}^\ast$).
Visualized for (a) full-precision, (b) PTQ-quantized I-BERT, (c) our method, and (d) QAT-quantized I-BERT, with token sub-sampling applied. 
Both (b) and (d) use i-GELU~\cite{i-bert} of the QAT-based approximation. (b) shows massive imbalance, highlighting the limitation of applying QAT-designed methods to PTQ settings in vision tasks. More results are presented in Appendix Fig~\textcolor{wacvblue}{4}.
}
\label{fig:ibert_gelu_output_dist}
\end{figure*}

\section{Motivation}
\label{sec:motivation}

To achieve integer-only vision transformers, approximation functions for all non-linear operations such as Softmax, GELU, and LayerNorm are essential. Existing QAT methods~\cite{mixed_v1, i-vit, i-bert} reduce approximation errors through retraining; however, these are dependent on training resources and hard to generalize to diverse model architectures. 
They also require full training datasets and involve complex hyperparameter tuning, which limits their applicability in real-world deployments.
Conversely, PTQ methods enable quantization without retraining but fail to fully quantize non-linear operations to integer operations, resulting in accuracy degradation.
An alternative approach is to directly use QAT-based approximation functions~\cite{i-bert, i-vit} in PTQ setting as a heuristic solution to the limitations of prior quantization methods.
However, as shown in Tab.~\ref{tab:motivation_qat2ptq_acc}, these heuristic applications of QAT-based functions (denoted as I-ViT${}^\ast$ and I-BERT${}^\ast$) result in a severe drop in accuracy to 0.08\% under both W8A8 and W4A8 settings. 
This performance loss is due to two key issues: (1) existing approximation functions are tailored for language data distributions and do not generalize well to vision tasks, and (2) PTQ performs quantization without retraining, making it difficult to compensate for non-linear function approximation errors.

Fig.~\ref{fig:ibert_gelu_output_dist} visualizes the distorted activation distributions of the GELU layer in ViT-B when employing I-BERT’s approximation function. 
Compared to Fig.~\ref{fig:ibert_gelu_output_dist} (a), Fig.~\ref{fig:ibert_gelu_output_dist} (b) demonstrates that an approximation function designed for language models fails to sustain a stable distribution under PTQ, leading to increased quantization error and significant performance degradation.
In contrast, Fig.~\ref{fig:ibert_gelu_output_dist} (d) shows that retraining can stabilize the activation distribution even with language-model-based approximations.
Additionally, Tab.~\ref{tab:motivation_qat2ptq_sensitivity} shows an analysis of quantization sensitivity in Signal-to-Quantization Noise Ratio (SQNR) for the GELU and Softmax layers of ViT-B.
A lower SQNR indicates higher quantization errors, meaning the layer is more sensitive to quantization and leads to accuracy loss.
The sensitivity dramatically increases in deeper layers for heuristic application of QAT-based functions (I-BERT${}^\ast$ and I-ViT${}^\ast$). 
Notably, I-ViT${}^\ast$ reports a $3.2 \times$ increase in quantization sensitivity at the 5-th block compared to the 4-th block.
Such increased sensitivities observed in both methods indicate that their simple approximation functions fail to reduce quantization errors in PTQ, leading to accuracy degradation.

To address these limitations, we propose approximation methods and a quantization pipeline that enables integer-only ViTs without retraining. Our IPTQ-ViT outperforms FQ-ViT, a state-of-the-art PTQ approach for non-linear operations, achieving higher accuracy under both W8A8 and W4A8 settings, as shown in Tab.~\ref{tab:motivation_qat2ptq_acc}.

\begin{table}[h]
\centering
\resizebox{\columnwidth}{!}{%
\begin{tabular}{lccccccc} 
\toprule
\textbf{Method} & \textbf{W/A} & \textbf{DeiT-T} & \textbf{DeiT-S} & \textbf{DeiT-B} & \textbf{ViT-B} & 
\textbf{Swin-T} & \textbf{Swin-S} \\
\midrule
Baseline & FP & 72.21 & 79.85 & 81.85 & 84.53 & 
81.35 & 83.20 \\
\midrule

\multirow{2}{*}{I-ViT${}^\ast$} 
& 8/8 & 61.66 & 49.65 & \textbf{0.10} & \textbf{0.10} & 
59.39 & \textbf{0.10} \\
& 4/8 & 58.98 & 56.44 & \textbf{0.33} & \textbf{0.34} & 
64.93 & \textbf{0.39} \\
\midrule

\multirow{2}{*}{I-BERT${}^\ast$}
& 8/8 & 68.37 & 77.31 & 80.88 & 81.71 & 
35 & 82.49 \\
& 4/8 & \textbf{0.08} & \textbf{0.10} & \textbf{0.10} & \textbf{0.09} & 
\textbf{0.10} & \textbf{0.10} \\
\midrule
\multirow{2}{*}{FQ-ViT~\cite{fq-vit}} 
& 8/8 & 71.61 & 79.17 & 81.20 & 83.31 & 
81.29 & 82.13 \\
& 4/8 & 66.91 & 76.93 & 79.99 & 78.73 & 
80.73 & 81.67 \\
\midrule

\multirow{2}{*}{IPTQ-ViT} 
& 8/8 & 72.10 & 79.76 & 81.84 & 84.19 & 
81.09 & 83.08 \\
& 4/8 & 66.90 & 77.28 & 80.98 & 82.03 & 
79.13 & 82.53 \\

\bottomrule
\end{tabular}%
}
\caption{Top-1 accuracy (\%) of QAT-based non-linear methods under PTQ on vision transformers (ImageNet-1k). ${}^\ast$ indicates direct PTQ application of QAT methods (I-ViT~\cite{i-vit}, I-BERT~\cite{i-bert}) using official code. "W/A" denotes weight/activation bit-width.}
\label{tab:motivation_qat2ptq_acc}
\end{table}

\begin{table}[h]
\centering
\scriptsize
\resizebox{\columnwidth}{!}{%
\begin{tabular}{lccccccc}
\toprule
\textbf{Method} & \textbf{Model} & \textbf{blk1} & \textbf{blk2} & \textbf{blk3} & \textbf{blk4} & \textbf{blk5} & \textbf{blk6} \\
\midrule
I-ViT${}^\ast$ & ViT-B & 0.50 & -1.10 & -4.15 & \textbf{-7.50} & \textbf{-24.13} & -32.66 \\
I-BERT${}^\ast$ & ViT-B & -11.25 & -16.00 & -11.44 & -12.13 & -12.40 & -12.47 \\
\midrule
\textbf{Method} & \textbf{Model} & \textbf{blk6} & \textbf{blk7} & \textbf{blk8} & \textbf{blk9} & \textbf{blk10} & \textbf{blk11} \\
\midrule
I-ViT${}^\ast$  & ViT-B & -34.65 & -27.42 & -24.61 & -21.04 & -17.22 & -10.72 \\
I-BERT${}^\ast$ & ViT-B & -11.35 & -8.98  & -13.43 & -22.62 & -25.08 & -17.81 \\
\bottomrule
\end{tabular}
}
\caption{Layer-wise quantization sensitivity (SQNR in dB, $\uparrow$ better) of QAT-based approximation functions on \emph{ViT-B} under PTQ. I-ViT${}^\ast$ applies Shiftmax~\cite{i-vit} to Softmax, and I-BERT${}^\ast$ uses i-GELU~\cite{i-bert} for GELU.}
\label{tab:motivation_qat2ptq_sensitivity}
\end{table}

\section{Method}
\label{sec:method}

\begin{figure*}[t]
\centering
\includegraphics[scale=0.33]{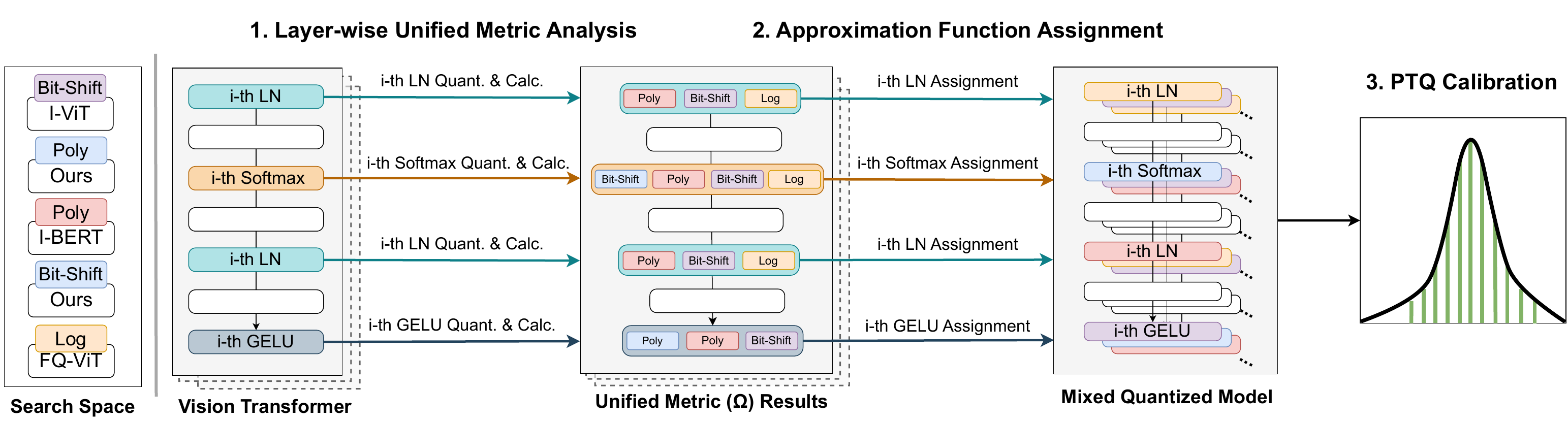}
\caption{Overview of IPTQ-ViT pipeline. In stage 1, each non-linear layer is quantized with all candidate approximation functions and the Unified Metric is computed for each case. Stage 2 assigns an approximation function that has a maximum metric value per activation layer. Stage 3 calibrates the mixed quantized model.}
\label{fig:overal_pipeline}
\centering
\end{figure*}

\subsection{Background}
\label{sec:method_preliminaries}
A quantizer is defined to map a real-valued input \( X \in \mathbb{R} \) to an integer value \( q \) within the range of \( b \)-bit integers, \(q = \operatorname{clip}(\lfloor X/s \rceil + z,\, 0,\, 2^b-1)\), 
where the scale \(s = (\alpha-\beta)/(2^b-1)\) and zero-point 
\(z = \operatorname{clip}(\lfloor -\beta/s \rceil,\, 0,\, 2^b-1)\). The quantization function is applied across the entire model. 
All linear (\textit{Conv}, \textit{Linear}, \textit{MatMul}) 
and non-linear (\textit{Softmax}, \textit{GELU}, \textit{LayerNorm}) 
operations are thus executed with integer arithmetic, enabling 
integer-only vision transformers.

\subsection{Overall Pipeline}
\label{sec:method_overal_pipeline}

Fig.~\ref{fig:overal_pipeline} illustrates the overall quantization pipeline of IPTQ-ViT. The framework consists of three stages: 
(1) conducting layer-wise \textit{Unified Metric} analysis, 
(2) assigning approximation functions to each non-linear layer based on the metric, 
(3) calibrating the mixed quantized model.

The search space of approximation functions is defined using four types: bit-shifting~\cite{i-vit}, polynomial~\cite{i-bert}, logarithm~\cite{fq-vit} and our proposed methods.
The supported search space for approximation functions varies by non-linear layer type:
(i) Softmax — logarithm, polynomial, bit-shifting, and our \textit{Efficient Bit-Softmax};  
(ii) GELU — polynomial, bit-shifting, and our \textit{Data-aware Poly-GELU};  
(iii) LayerNorm — logarithm, polynomial, and bit-shifting.
For each non-linear layer, we compute a metric for all candidate approximation functions and assign the one with the highest score. This layer-wise selection forms a mixed-quantized model.
For instance, ViT-B contains 49 activation layers: 2 LayerNorms, 1 Softmax, and 1 GELU per Transformer block (12 blocks), plus one additional LayerNorm before the first block. Thus, the full model can be constructed with 159 computations. Finally, PTQ calibration is applied to this model to finalize quantization parameters.

\subsection{Data-aware Poly-GELU for Integer-only \newline GELU}
\label{sec:method_gelu}

The proposed \textit{Data-aware Poly-GELU} is a polynomial-based GELU approximation optimized for vision tasks in PTQ to support integer-only inference. It addresses the limitations of i-GELU~\cite{i-bert}, originally designed for language models in I-BERT~\cite{i-bert}, as discussed in Section~\ref{sec:motivation}.
{\small
 \begin{align}
    \text{GELU}(x) = \frac{1}{2} \cdot x \cdot \left(1 + \text{erf} \left( \frac{x}{\sqrt{2}} \right) \right) & \label{eq:gelu}
\end{align}
}

I-BERT~\cite{i-bert} approximates the error function (erf) with a second-order polynomial, denoted as $L_{ibert}$, to implement integer-only GELU inference. In Eq.~(\ref{eq:ibert-gelu-opti}), i-GELU~\cite{i-bert} directly solves an optimization problem targeting the GELU function defined in Eq.~(\ref{eq:gelu}). However, this approach introduces complexity due to multiplicative terms: $x$, $1/2$, and $1 + \text{erf}(\frac{x}{\sqrt{2}})$, making the optimization sensitive to input distribution.
To solve the problem, we simplify the optimization problem of I-BERT~\cite{i-bert} by formulating the approximation as an optimization over its core component, the error function (erf), as shown in Eq.~(\ref{eq:ours-gelu-opti}). This simplification allows for a more stable and accurate approximation of erf. 
{\small
 \begin{align}
    \min_{a, b, c} \frac{1}{2} \left\lVert \text{GELU}(x) - x \cdot \frac{1}{2} \left[ 1 +  L_{ibert}\left( \frac{x}{\sqrt{2}} \right) \right] \right\rVert_2^2 & \label{eq:ibert-gelu-opti} \\
    \text{s.t.} \quad L_{ibert}(x) = a(x + b)^2 + c & \nonumber
\end{align}
}

In contrast to the fixed approximation range based on language data used by the previous method~\cite{i-bert}, 
our approach determines the approximation range by calculating the minimum and maximum values from the vision data and recalculates the polynomial coefficients accordingly. 
Further details on the approximation range and its effect can be found in Appendix~\textcolor{wacvblue}{G.1}.
Additionally, we extend a quartic polynomial to enhance accuracy for erf approximation, defined in Eq.~(\ref{eq:ours-erf}), with coefficients $a$ is $-0.019913$ and $b$ is $-2.698088$. These coefficients are pre-quantized as a constant before inference.
{\small
\begin{eqnarray} 
L_{\text{ours}}(x) = \text{sign}(x) \cdot \left[ a \cdot \left( \text{clip}(|x|, \text{max}=-b) + b \right) \right]^4 + 1 \label{eq:ours-erf} \\
\arg\min_{a, b} \sum_{i=1}^{N} \left\|  \text{erf}(x_i) - L_{\text{ours}}(x_i; a, b) \right\|_2^2 \label{eq:ours-gelu-opti} \\ 
\text{for } x \in [\min(T), \max(T)] \nonumber 
\end{eqnarray}
}

The resulting polynomial GELU function is defined in Eq.~(\ref{eq:ours-gelu}). 
\textit{Data-aware Poly-GELU} establishes a trade-off between computational cost and accuracy depending on the polynomial degree. We empirically determine the optimal polynomial degree by evaluating model accuracy across different degrees within our quantization pipeline.

{\small
    \begin{equation}
    \text{Data-aware-Poly-GELU}(x) = \frac{1}{2} \cdot x \cdot 
    \left[1 + L_{\text{ours}} \left( \frac{x}{\sqrt{2}} \right) \right]
    \label{eq:ours-gelu}
    \end{equation}
}

\begin{figure}[t]
\includegraphics[width=0.48\textwidth]{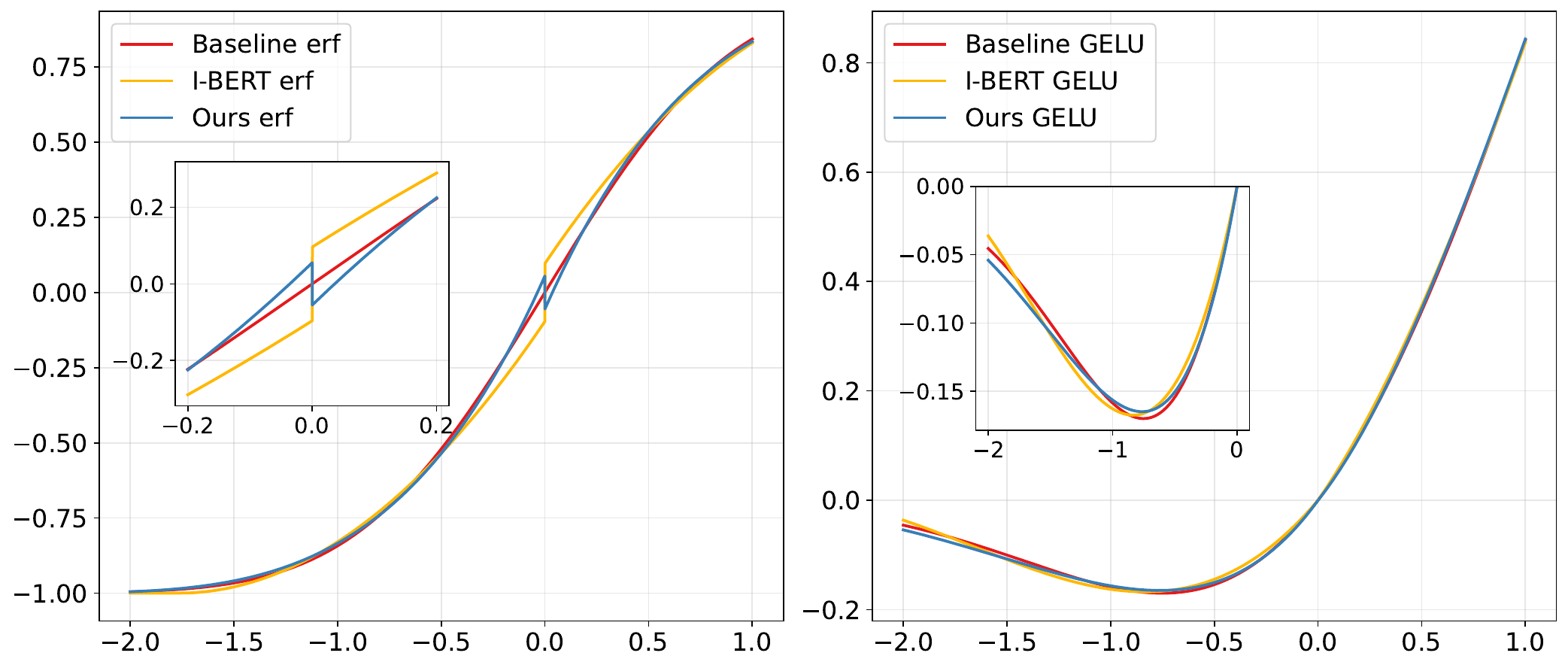}
\caption{Left: Comparison of \textit{erf} approximations by baseline, I-BERT~\cite{i-bert}, and ours. Right: Comparison of GELU approximations by baseline, i-GELU~\cite{i-bert}, and ours.}
\label{fig:abliation_gelu_graph}
\small
\end{figure}

We evaluated our polynomial GELU approximation (Eq.~(\ref{eq:ours-gelu})) with varying polynomial degrees on ImageNet-1k using our quantization pipeline. As the degree increases, accuracy consistently improves: 79.24\%, 79.30\%, and 79.76\% for degrees 2, 3, and 4 on DeiT-S, and 81.22\%, 81.30\%, and 81.84\% on DeiT-B, respectively. Although a quartic polynomial incurs higher computational cost, we reduce this overhead by reusing integer and squared terms. Given its accuracy gain, we select the quartic polynomial.
Fig.~\ref{fig:abliation_gelu_graph} illustrates our proposed methods for erf and GELU, showing their close similarity to the original functions. As shown in Tab.~\ref{tab:erf_gelu_approximation_error}, both the $L^2$ and $L^\infty$ approximation errors of our proposed method are reduced compared to previous methods~\cite{i-bert}.

\begin{table}[h]
\footnotesize
\centering
\begin{tabular}{lcc}
\toprule
\textbf{Method} & \textbf{$L^2$ error} & \textbf{$L^\infty$ error} \\
\midrule
erf I-BERT~\cite{i-bert} & 0.0264 & 0.0962 \\
erf Eq.~(\ref{eq:ours-erf}) (Ours) & 0.0098 & 0.0550 \\
\midrule
i-GELU~\cite{i-bert} & 0.0094 & 0.0182 \\
$\text{Data-aware Poly-GELU}$ (Ours) & 0.0051 & 0.0093 \\
\bottomrule
\end{tabular}
\caption{Comparison of our method and I-BERT~\cite{i-bert} for approximating \textit{erf} and GELU. $L^2$ and $L^\infty$ errors are computed over the range $(-3, 3)$.}
\label{tab:erf_gelu_approximation_error}
\end{table}

\subsection{Efficient Bit-exp for Integer-only Softmax} \label{sec:method_bit_exp}

Softmax converts input values into a probability distribution, representing the relative importance of each element. To prevent overflow of Softmax approximation, previous methods~\cite{i-bert, fq-vit, i-vit} generally subtract the maximum value from the input of the exponential function. We apply the same stabilization in Eq.~(\ref{eq:softmax_numerical_stability}).
{\small
\begin{eqnarray} \label{eq:softmax_numerical_stability}
\text{Softmax}(x_i) = 
\frac{e^{S_{x_i} \cdot (Q_{x_i} - Q_{\text{max}})}}{\sum\limits_{j=1}^{d} e^{S_{x_j} \cdot (Q_{x_j} - Q_{\text{max}})}} =
\frac{e^{S_{\Delta_i} \cdot Q_{\Delta_i}}}{\sum\limits_{j=1}^{d} e^{S_{\Delta_j} \cdot Q_{\Delta_j}}} 
\end{eqnarray}
}

In Section~\ref{sec:motivation}, we identified significant accuracy degradation when directly applying I-ViT~\cite{i-vit} to PTQ. Here, we highlight the overly simple bit-shift approximation (Shiftmax) used in I-ViT~\cite{i-vit} as a primary cause. Firstly, we summarize the Shiftmax~\cite{i-vit} approximation process. To approximate Softmax, one must approximate the exponential function, whose input is expressed as a product of scaling factor $S_{\Delta}$ and quantized tensor $Q_{\Delta}$. The exponential function is reformulated into base-2 exponential form, as shown in Eq.~(\ref{eq:ivit_stage1}). 
In Eq.~(\ref{eq:ivit_stage2}), $S_{\Delta} \cdot Q_{\Delta}$ is decomposed into an integer component $-q$ and a fractional component $S_{\Delta} \cdot (-r)$. An integer component $-q$ is represented using bit-shift operations, as defined in Eq.~(\ref{eq:ivit_stage3}). I-ViT~\cite{i-vit} further approximates the left-hand side of Eq.~(\ref{eq:ivit_stage4}) by a linear function in the form of $1 + x/2$.
{\small
    \begin{align}
    e^{S_{\Delta} \cdot Q_{\Delta}} 
    &= 2^{S_{\Delta} \cdot Q_{\Delta} \cdot \log_2 e} \nonumber \\
    &\approx 2^{S_{\Delta} \cdot (Q_{\Delta} + (Q_{\Delta} \gg 1) - (Q_{\Delta} \gg 4))} 
    = 2^{S_{\Delta} \cdot Q_p} \label{eq:ivit_stage1} \\
    2^{S_{\Delta} \cdot Q_p}
    &= 2^{-q + S_{\Delta} \cdot (-r)} \label{eq:ivit_stage2} \\
    &= 2^{S_{\Delta} \cdot (-r)} \gg q \label{eq:ivit_stage3} \\
    2^{S_{\Delta} \cdot (-r)} 
    &\approx \left[ S_{\Delta} \cdot (-r) \right] / 2 + 1 \label{eq:ivit_stage4}
    \end{align}
}

While retraining in QAT can compensate for quantization errors introduced by bit-shift approximations, it is not feasible in PTQ. To address the limitation, we propose an approximation function, \textit{Efficient Bit-exp}. 
We reformulate the left-hand side of Eq.~(\ref{eq:ivit_stage4}) as a base-2 exponential function, replacing the overly simple linear form. It is then approximated with a Taylor series, as defined in Eq.~(\ref{eq:softmax_expansion}). High-order polynomials increase computational cost. 
Given the limited range of fractional inputs $S_{\Delta} \cdot (-r)$, we adopt a first-degree polynomial (Eq.~(\ref{eq:softmax_degree1})) to minimize computational overhead. 
Furthermore, we approximate the constant $ln 2$ as the binary value $(0.1011)_b$, allowing us to implement the exponential function solely with bit-shift and addition operations, eliminating complex multiplications. Our improved exponential approximation is defined in Eq.~(\ref{eq:ours-exp}). 
{\small
    \begin{align}
    2^X &= e^{\ln{2} \cdot X } \approx \sum_{d=0}^{D} \frac{(\ln{2} \cdot X )^d}{d!} \label{eq:softmax_expansion} \\
    &\approx 1 + \ln{2} \cdot X \quad \text{for } D = 1 \label{eq:softmax_degree1} \\
    2^{S_{\Delta} \cdot (-r)} &\approx \left[  \Phi(S_{\Delta} \cdot (-r)) \right] + 1 \label{eq:softmax_taylor_ours} \\
    \text{s.t.} \quad \Phi(x) &= x \gg 1 + x \gg 3 + x \gg 4 \nonumber \\
    \text{Efficient-Bit-exp}(x) &= S_{\Delta} \cdot \left[ \Phi(-r)  + \left\lfloor1/S_{\Delta} \right\rfloor \right] \label{eq:ours-exp}
    \end{align}
}
\textit{Efficient Bit-exp} approximates the numerator of Eq.~(\ref{eq:softmax_numerical_stability}), while the denominator is represented by the sum of these approximations. To generate the probability distribution, we adopt integer division (IntDiv) introduced by I-ViT~\cite{i-vit}.  
Based on \textit{Efficient Bit-exp} and IntDiv, we define our integer-only Softmax function, \textit{Efficient Bit-Softmax}, as shown in Eq.~(\ref{eq:softmax_ours_final}), where $M$ is a large integer to prevent overflow and $b$ denotes the bit-width. This precise exponential approximation enables integer-only Softmax computation under PTQ.
{\small
\begin{align}
\label{eq:softmax_ours_final}
\text{Efficient-Bit-Softmax}(x) 
&= \frac{Q_{\exp_i}}{\sum_{j}^{d} Q_{\exp_j}} \nonumber \\
&= \text{IntDiv} \left( Q_{\exp_i}, \sum_{j}^{d} Q_{\exp_j}, b \right) \nonumber \\
&= \left( \left\lfloor \frac{2^M}{\sum_{j}^{d} Q_{\exp_j}} \right\rfloor \cdot Q_{\exp_i} \right) \nonumber \\
&\quad \gg \left( M - (b - 1) \right) \\
\text{s.t.} \quad Q_{\exp_i} 
&= \text{Efficient-Bit-exp}(Q_i)  \nonumber
\end{align}
}
 
As shown in Tab.~\ref{tab:softmax_degree_ablation}, the first-degree approximation ("Ours-D1") of Eq.~(\ref{eq:softmax_expansion}) achieves higher accuracy than the second-degree ("Ours-D2") across all evaluated models and bit-widths. Therefore, we choose a first-degree polynomial approximation to balance accuracy and efficiency. \news{A detailed analysis of the lower accuracy of the second-order approximation is provided in Appendix~\textcolor{wacvblue}{H}.}
Tab.~\ref{tab:softmax_2x_error} presents a comparison of the $L^2$ and $L^\infty$ approximation errors between our method and I-ViT~\cite{i-vit} in the fractional range $(-1, 1)$. Our method achieves higher approximation accuracy than I-ViT~\cite{i-vit}, facilitating an efficient integer-only implementation appropriate for PTQ.

\begin{table}[h]
\centering
\resizebox{\columnwidth}{!}{%
\begin{tabular}{lccccccc}
\toprule
\textbf{Method} & \textbf{W/A} & \textbf{DeiT-T} & \textbf{DeiT-S} & \textbf{DeiT-B} & 
\textbf{ViT-B} & 
\textbf{Swin-T} & \textbf{Swin-S} \\
\midrule
Ours-D1          & 8/8   & 72.10 & 79.76 & 81.84 & 84.19 & 
81.19 & 83.08 \\ 

Ours-D2          & 8/8   & 71.93 & 79.37 & 81.62 & 84.03 & 
80.90 & 82.95 \\

Ours-D1          & 4/8   & 66.90 & 77.28 & 80.98 & 82.03 & 
79.13 & 82.53 \\

Ours-D2          & 4/8   & 65.93 & 76.16 & 80.01 & 80.87 & 
\news{78.59} & 81.62 \\
\bottomrule
\end{tabular}%
}
\caption{Top-1 accuracy (\%) for different polynomial degrees of Eq.~(\ref{eq:softmax_ours_final}) on the ImageNet-1K. "D" denotes degree. Degree 1 shows better performance on W8A8 and W4A8 than degree 2.}
\label{tab:softmax_degree_ablation}
\end{table}

\begin{table}[h]
\centering
\footnotesize
\begin{tabular}{lcc}
\toprule
\textbf{Method} & \textbf{$L^2$ error} & \textbf{$L^\infty$ error} \\
\midrule
base-2 exp (I-ViT~\cite{i-vit}) & 0.1717 & 0.5 \\
base-2 exp (Ours) & 0.1126 & 0.3069 \\
\bottomrule
\end{tabular}
\label{tab:error_comparison}
\caption{Comparison of base-2 exponential approximation functions. $L^2$ and $L^\infty$ errors are evaluated over the range $(-1, 1)$.}
\label{tab:softmax_2x_error}
\end{table}

\begin{table*}[t]
\centering
\resizebox{1.9\columnwidth}{!}{
\begin{tabular}{l|c|c|c|cc|cc|cc|cc|cc|cc}
\toprule
\textbf{Method} & \textbf{Type} & \textbf{Opt} & \textbf{W/A} &
\textbf{DeiT-T} & \textbf{Diff} &
\textbf{DeiT-S} & \textbf{Diff} &
\textbf{DeiT-B} & \textbf{Diff} &
\textbf{ViT-B} & \textbf{Diff} &
\textbf{Swin-T} & \textbf{Diff} &
\textbf{Swin-S} & \textbf{Diff} \\
\midrule
Baseline & FP & FP & FP &
72.21 & - & 79.85 & - & 81.85 & - & 84.53 & - & 81.35 & - & 83.20 & - \\
\midrule
I-BERT~\cite{i-bert} & \multirow{2}{*}{QAT} & IO & \multirow{2}{*}{8/8} &
71.33 & \textcolor{ForestGreen}{\textbf{+0.77}} &
79.11 & \textcolor{ForestGreen}{\textbf{+0.65}} &
80.79 & \textcolor{ForestGreen}{\textbf{+1.05}} &
83.70 & \textcolor{ForestGreen}{\textbf{+0.49}} &
80.15 & \textcolor{ForestGreen}{\textbf{+0.94}} &
81.86 & \textcolor{ForestGreen}{\textbf{+1.22}} \\
I-ViT~\cite{i-vit} &  & IO &  &
72.24 & \textcolor{BrickRed}{-0.14} &
80.12 & \textcolor{BrickRed}{-0.36} &
81.74 & \textcolor{ForestGreen}{\textbf{+0.10}} &
84.76 & \textcolor{BrickRed}{-0.57} &
81.50 & \textcolor{BrickRed}{-0.41} &
83.01 & \textcolor{ForestGreen}{\textbf{+0.07}} \\
\midrule
I-ViT${}^{\ast}$ & \multirow{10}{*}{PTQ} & IO & \multirow{10}{*}{8/8} &
61.66 & \textcolor{ForestGreen}{\textbf{+10.44}} &
49.65 & \textcolor{ForestGreen}{\textbf{+30.11}} &
0.10 & \textcolor{ForestGreen}{\textbf{+81.74}} &
0.10 & \textcolor{ForestGreen}{\textbf{+84.09}} &
59.39 & \textcolor{ForestGreen}{\textbf{+21.70}} &
0.10 & \textcolor{ForestGreen}{\textbf{+82.98}} \\
I-BERT${}^{\ast}$ &  & IO &  &
68.37 & \textcolor{ForestGreen}{\textbf{+3.73}} &
77.31 & \textcolor{ForestGreen}{\textbf{+2.44}} &
80.88 & \textcolor{ForestGreen}{\textbf{+0.96}} &
81.71 & \textcolor{ForestGreen}{\textbf{+2.48}} &
35.00 & \textcolor{ForestGreen}{\textbf{+46.09}} &
82.49 & \textcolor{ForestGreen}{\textbf{+0.59}} \\
FQ-ViT~\cite{fq-vit} &   & PF  &   &
71.61 & \textcolor{ForestGreen}{\textbf{+0.49}} &
79.17 & \textcolor{ForestGreen}{\textbf{+0.59}} &
81.20 & \textcolor{ForestGreen}{\textbf{+0.64}} &
83.31 & \textcolor{ForestGreen}{\textbf{+0.88}} &
81.29 & \textcolor{BrickRed}{-0.20} &
82.13 & \textcolor{ForestGreen}{\textbf{+0.95}} \\
PTQ4ViT~\cite{ptq4vit} &   & PF  &   &
71.72 & \textcolor{ForestGreen}{\textbf{+0.38}} &
79.47 & \textcolor{ForestGreen}{\textbf{+0.29}} &
81.48 & \textcolor{ForestGreen}{\textbf{+0.36}} &
84.25 & \textcolor{BrickRed}{-0.06} &
81.24 & \textcolor{BrickRed}{-0.15} &
83.10 & \textcolor{BrickRed}{-0.02} \\
RepQ-ViT~\cite{repq-vit} &   & PF  &   &
72.05 & \textcolor{ForestGreen}{\textbf{+0.05}} &
79.55 & \textcolor{ForestGreen}{\textbf{+0.21}} &
81.45 & \textcolor{ForestGreen}{\textbf{+0.39}} &
81.45 & \textcolor{ForestGreen}{\textbf{+2.74}} &
81.28 & \textcolor{BrickRed}{-0.19} &
82.34 & \textcolor{ForestGreen}{\textbf{+0.74}} \\
PSAQ-ViT V1~\cite{psaq-vit-v1} &   & PF  &   &
71.56 & \textcolor{ForestGreen}{\textbf{+0.54}} &
76.92 & \textcolor{ForestGreen}{\textbf{+2.84}} &
79.10 & \textcolor{ForestGreen}{\textbf{+2.74}} &
37.36 & \textcolor{ForestGreen}{\textbf{+46.83}} &
75.35 & \textcolor{ForestGreen}{\textbf{+5.74}} &
76.64 & \textcolor{ForestGreen}{\textbf{+6.44}} \\
PSAQ-ViT V2~\cite{psaq-vit-v2} &  & PF   &   &
72.17 & \textcolor{BrickRed}{-0.07} &
79.56 & \textcolor{ForestGreen}{\textbf{+0.20}} &
81.52 & \textcolor{ForestGreen}{\textbf{+0.32}} &
N/A & N/A &
80.20 & \textcolor{ForestGreen}{\textbf{+0.89}} &
82.13 & \textcolor{ForestGreen}{\textbf{+0.95}} \\
NoisyQuant-Linear~\cite{noisyquant} &  & PF &   &
N/A & N/A &
79.11 & \textcolor{ForestGreen}{\textbf{+0.65}} &
81.30 & \textcolor{ForestGreen}{\textbf{+0.54}} &
84.10 & \textcolor{ForestGreen}{\textbf{+0.09}} &
81.05 & \textcolor{ForestGreen}{\textbf{+0.04}} &
83.07 & \textcolor{ForestGreen}{\textbf{+0.01}} \\
NoisyQuant-PTQ4ViT~\cite{noisyquant} &  & PF &   &
N/A & N/A &
79.51 & \textcolor{ForestGreen}{\textbf{+0.25}} &
81.45 & \textcolor{ForestGreen}{\textbf{+0.39}} &
84.22 & \textcolor{BrickRed}{-0.03} &
81.25 & \textcolor{BrickRed}{-0.16} &
83.13 & \textcolor{BrickRed}{-0.05} \\
CLAMP-ViT~\cite{clamp-vit} &  & PF &   &
72.17 & \textcolor{BrickRed}{-0.07} &
79.55 & \textcolor{ForestGreen}{\textbf{+0.21}} &
81.77 & \textcolor{ForestGreen}{\textbf{+0.07}} &
84.19 & \textcolor{ForestGreen}{\textbf{+0.00}} &
81.17 & \textcolor{BrickRed}{-0.08} &
82.50 & \textcolor{ForestGreen}{\textbf{+0.58}} \\
\midrule
IPTQ-ViT & PTQ & IO & 8/8 &
\multicolumn{2}{c|}{72.10} &
\multicolumn{2}{c|}{79.76} &
\multicolumn{2}{c|}{81.84} &
\multicolumn{2}{c|}{84.19} &
\multicolumn{2}{c|}{81.09} &
\multicolumn{2}{c|}{83.08} \\
\midrule
FQ-ViT~\cite{fq-vit} & \multirow{9}{*}{PTQ} & PF &  \multirow{9}{*}{4/8} &
66.91 & \textcolor{BrickRed}{-0.01} &
76.93 & \textcolor{ForestGreen}{\textbf{+0.35}} &
79.99 & \textcolor{ForestGreen}{\textbf{+0.99}} &
78.73 & \textcolor{ForestGreen}{\textbf{+3.30}} &
80.73 & \textcolor{BrickRed}{-1.60} &
81.67 & \textcolor{ForestGreen}{\textbf{+0.86}} \\
I-ViT${}^{\ast}$ &  & IO &   &
58.98 & \textcolor{ForestGreen}{\textbf{+7.92}} &
56.44 & \textcolor{ForestGreen}{\textbf{+20.84}} &
0.33 & \textcolor{ForestGreen}{\textbf{+80.65}} &
0.34 & \textcolor{ForestGreen}{\textbf{+81.69}} &
64.93 & \textcolor{ForestGreen}{\textbf{+14.20}} &
0.39 & \textcolor{ForestGreen}{\textbf{+82.14}} \\
I-BERT${}^{\ast}$ &  & IO &   &
0.08 & \textcolor{ForestGreen}{\textbf{+66.82}} &
0.10 & \textcolor{ForestGreen}{\textbf{+77.18}} &
0.10 & \textcolor{ForestGreen}{\textbf{+80.88}} &
0.09 & \textcolor{ForestGreen}{\textbf{+81.94}} &
0.10 & \textcolor{ForestGreen}{\textbf{+79.03}} &
0.10 & \textcolor{ForestGreen}{\textbf{+82.43}} \\
PTQ4ViT~\cite{ptq4vit} &  & PF &   &
66.57 & \textcolor{ForestGreen}{\textbf{+0.33}} &
76.96 & \textcolor{ForestGreen}{\textbf{+0.32}} &
79.47 & \textcolor{ForestGreen}{\textbf{+1.51}} &
67.99 & \textcolor{ForestGreen}{\textbf{+14.04}} &
N/A & N/A &
79.62 & \textcolor{ForestGreen}{\textbf{+2.91}} \\
RepQ-ViT~\cite{repq-vit} &  & PF &   &
68.75 & \textcolor{BrickRed}{-1.85} &
76.75 & \textcolor{ForestGreen}{\textbf{+0.53}} &
80.12 & \textcolor{ForestGreen}{\textbf{+0.86}} &
76.29 & \textcolor{ForestGreen}{\textbf{+5.74}} &
80.51 & \textcolor{BrickRed}{-1.38} &
82.14 & \textcolor{ForestGreen}{\textbf{+0.39}} \\
PSAQ-ViT V1~\cite{psaq-vit-v1} &  & PF &   &
65.57 & \textcolor{ForestGreen}{\textbf{+1.33}} &
73.23 & \textcolor{ForestGreen}{\textbf{+4.05}} &
77.05 & \textcolor{ForestGreen}{\textbf{+3.93}} &
25.34 & \textcolor{ForestGreen}{\textbf{+56.69}} &
71.79 & \textcolor{ForestGreen}{\textbf{+7.34}} &
75.14 & \textcolor{ForestGreen}{\textbf{+7.39}} \\
PSAQ-ViT V2~\cite{psaq-vit-v2} &   & PF &   &
68.61 & \textcolor{BrickRed}{-1.71} &
76.36 & \textcolor{ForestGreen}{\textbf{+0.92}} &
79.49 & \textcolor{ForestGreen}{\textbf{+1.49}} &
N/A & N/A &
76.28 & \textcolor{ForestGreen}{\textbf{+2.85}} &
78.86 & \textcolor{ForestGreen}{\textbf{+3.67}} \\
CLAMP-ViT~\cite{clamp-vit} &   & PF &   &
69.93 & \textcolor{BrickRed}{-3.03} &
77.03 & \textcolor{ForestGreen}{\textbf{+0.25}} &
80.93 & \textcolor{ForestGreen}{\textbf{+0.05}} &
78.73 & \textcolor{ForestGreen}{\textbf{+3.30}} &
80.28 & \textcolor{BrickRed}{-1.15} &
82.51 & \textcolor{ForestGreen}{\textbf{+0.02}} \\
\midrule
IPTQ-ViT & PTQ & IO & 4/8 &
\multicolumn{2}{c|}{66.90} & 
\multicolumn{2}{c|}{77.28} &
\multicolumn{2}{c|}{80.98} &
\multicolumn{2}{c|}{82.03} &
\multicolumn{2}{c|}{79.13} &
\multicolumn{2}{c|}{82.53} \\
\bottomrule
\end{tabular}
}

\caption{Top-1 accuracy (\%) of quantized ViTs on image classification with the ImageNet-1k. ${}^{\ast}$ denotes results reproduced with the official code. "Diff" denotes the accuracy gap between IPTQ-ViT and compared methods. \textbf{\textcolor{ForestGreen}{Green}} and \textbf{bold} indicate IPTQ-ViT improvements. "IO": all ops computed in integer arithmetic. "PF": at least one op executed in floating point.}
\label{tab:all_results_w4a8_w8a8}
\end{table*}

\subsection{Unified Metric for Approximation Function Assignment}
\label{sec:method_alloc_metric}

In order to assign an optimal non-linear approximation function to each activation layer in PTQ, we propose a \textit{Unified Metric} that jointly considers three factors: quantization sensitivity ($\mathcal{Q}$), quantization perturbation ($\mathcal{P}$), and computational cost ($\mathcal{C}$). The approximation function with the highest \textit{Unified Metric} score is selected for each layer during stage 2 of the pipeline (Fig.~\ref{fig:overal_pipeline}).

Although SQNR is widely used to assess quantization sensitivity, it measures relative error on a logarithmic scale and may obscure meaningful differences in absolute error.
For example, in Eq.~\ref{eq:sqnr}, assuming an input power of $\mathbb{E}[(X)^2] = 10^4$, perturbations of 100 and 60 correspond to SQNR values of 20 dB and 22.21 dB, respectively. While the SQNR difference is only 2.21 dB, the absolute error difference is 40, which may significantly affect accuracy in deeper layers.
This suggests that SQNR alone may not adequately capture error accumulation.
To address this limitation, we additionally incorporate $\mathcal{P}$, as defined in Eq.~(\ref{eq:perturbation}), jointly considering both SQNR and perturbation provides richer guidance for assigning appropriate approximation functions.

To estimate the efficiency of each approximation function, we include the number of arithmetic and bit operations in \textit{Unified Metric} ($\Omega$), counting as in FLOPs.
Similar to prior quantization works, we use this operation count as an indirect metric for computational cost. Since the factors of $\Omega$ differ in scales and signs, we apply the Softplus function as defined in Eq.~(\ref{eq:softplus}) to normalize components of the metric. The transformed $N(\mathcal{P})$ and $N(\mathcal{C})$ are converted to reciprocal forms, as lower values are preferable. As shown in Eq.~(\ref{eq:metric}), \textit{Unified Metric} is defined as the harmonic mean of the three factors to balance their contributions, which prevents disproportionate influence from extreme values.
{\small
    \begin{align}
        \Omega &= \sum^{L}_{i} \frac{3}{N(\mathcal{Q}_i)^{-1} + N(\mathcal{P}_i) + N(\mathcal{C}_i)} \label{eq:metric} \\
        N(x) &= \log \left(1 + \exp(x) \right) \label{eq:softplus} \\
        \mathcal{Q} &= 20 \log \frac{\mathbb{E}[X^2]}{\mathbb{E}[(X - Q)^2]} \label{eq:sqnr} \\
        \mathcal{P} &= \left\| X - Q \right\|_2^2 \label{eq:perturbation} \\
        \mathcal{C} &= \text{Integer operation count of the layer} \label{eq:complexity}
    \end{align}
}
\section{Experiments} 
\label{sec:experiments}
\subsection{Experimental Setup} 
\label{sec:experiments_setup}
We evaluate IPTQ-ViT on both image classification and object detection with W8A8 and W4A8 settings, where “W” and “A” denote weight and activation bit-width. We adopt symmetric, channel-wise quantization for weights and asymmetric, layer-wise quantization for activations. For comparison, we include PTQ methods~\cite{ptq4vit, repq-vit, psaq-vit-v1, psaq-vit-v2, noisyquant, clamp-vit} and integer-only QAT baselines (I-BERT~\cite{i-bert}, I-ViT~\cite{i-vit}), with their non-linear approximations re-implemented under PTQ (denoted ${}^\ast$). All experiments are conducted on a single NVIDIA RTX 3090 GPU.

\noindent\textbf{Image Classification.}
We benchmark ViT~\cite{vit}, DeiT~\cite{deit}, and Swin~\cite{swin} on ImageNet-1K~\cite{imagenet}, reporting top-1 accuracy against PTQ and integer-only QAT methods.

\noindent\textbf{Object Detection.}
We evaluate Swin-T and Swin-S backbones within Cascade Mask R-CNN~\cite{cai2018cascade} on COCO~\cite{lin2014microsoft} using MMDetection~\cite{chen1906mmdetection}, comparing box and mask AP with prior PTQ baselines.

\noindent\textbf{Latency.} 
We measure end-to-end latency of batch size 8 by deploying IPTQ-ViT with TVM~\cite{chen2018tvm} in W8A8. I-ViT~\cite{i-vit} is the only prior integer-only ViTs reporting latency, but its evaluation recipe (e.g., auto-tuning settings, warm-up iterations, and repetition counts) was not disclosed. To ensure a fair comparison, we evaluated I-ViT~\cite{i-vit}, the FP32 baseline, and our method under an identical evaluation setting. Full details are provided in Appendix~\textcolor{wacvblue}{J}.

\noindent\textbf{Calibration.}
We uses randomly sampled training images. Results under varying calibration set sizes are provided in Appendix~\textcolor{wacvblue}{C.3} and~\textcolor{wacvblue}{D.1}.

\begin{table*}[t]
\centering
\resizebox{1.9\columnwidth}{!}{%
\begin{tabular}{l|cccc|ccccc}
\toprule
\textbf{SearchSpace / Metric} & \textbf{GELU} & \textbf{Softmax} & \textbf{SQNR} & \textbf{Unified} &
\textbf{DeiT-T} & \textbf{DeiT-S} & \textbf{DeiT-B} & \textbf{Swin-T} & \textbf{Swin-S} \\
\midrule

\multicolumn{5}{c}{\textit{Baseline FQ-ViT~\cite{fq-vit}}} & 71.61 & 79.17 & 81.20 & 81.29 & 82.13 \\
\midrule

Legacy / SQNR &  &  & \checkmark &  & 71.48 & 78.83 & 80.99 & 78.96 & 82.27 \\
Legacy / Unified &  &  &  & \checkmark & 71.16 & 79.03 & 81.20 & 80.01 & 82.3 \\
\midrule

Extended (Softmax) / SQNR & & \checkmark & \checkmark &  & 72.08 & 79.52 & \textbf{81.67} & 80.09 & 82.37 \\
Extended (GELU) / SQNR & \checkmark &  & \checkmark &  & 71.93 & 79.50 & \textbf{81.28} & 80.17 & 82.43 \\
\midrule

Extended / SQNR & \checkmark & \checkmark & \checkmark &  & 71.97 & 79.66 & \textbf{81.73} & 80.42 & 82.57 \\
Extended / Unified & \checkmark & \checkmark &  & \checkmark & 72.10 & 79.76 & \textbf{81.84} & 81.09 & 83.08 \\

\bottomrule
\end{tabular}
}
\caption{
Ablation study on ImageNet-1K (top-1 accuracy, \%) under W8A8. 
\checkmark indicates the use of a component: \textit{Data-aware Poly-GELU} ("GELU"), \textit{Efficient Bit-Softmax} ("Softmax"), and \textit{Unified Metric} ("Unified"). 
"Legacy" denotes the search space with prior approximations (I-ViT~\cite{i-vit}, I-BERT~\cite{i-bert}, FQ-ViT~\cite{fq-vit}), and "Extended" augments it with our proposed methods. 
}
\label{tab:metric_ablation_summary}
\end{table*}

\subsection{Evaluation on Image Classification}
\label{sec:exp_img_cls}
As shown in Tab.~\ref{tab:all_results_w4a8_w8a8}, IPTQ-ViT consistently matches or has better accuracy at W8A8 and W4A8.
In W8A8, IPTQ-ViT achieves average accuracy gains of 0.66\%p on DeiT-S, 0.68\%p on DeiT-B, and 7.21\%p on Swin-S compared to other PTQ methods, except NoisyQuant-PTQ4ViT~\cite{noisyquant} on Swin-S. IPTQ-ViT surpasses FQ-ViT~\cite{fq-vit}, the SOTA PTQ method for fully quantized ViTs, on all models except Swin-T, while maintaining integer-only inference.
Against integer-only QAT methods~\cite{i-bert, i-vit}, IPTQ-ViT consistently outperforms I-BERT~\cite{i-bert} and exceeds I-ViT~\cite{i-vit} on DeiT-B and Swin-S, without any retraining.
These results demonstrate the effectiveness of our methods. A detailed ablation study is presented in Section~\ref{sec:ablation_preliminary}.
Moreover, directly applying QAT-based functions in PTQ (I-BERT${}^{\ast}$ and I-ViT${}^{\ast}$) leads to considerable accuracy degradation in both W8A8 and W4A8, highlighting the difficulty of approximating non-linear operations without retraining. 

\subsection{Evaluation on Object Detection}
\label{sec:exp_object_det}

In Tab.~\ref{tab:swin_results_od}, IPTQ-ViT delivers competitive performance across Swin-T and Swin-S backbones under various quantization settings. Notably, with Swin-S at W8A8, we surpass FQ-ViT~\cite{fq-vit} by +1.0 box AP and +0.7 mask AP. Compared to RepQ-ViT~\cite{repq-vit} and CLAMP-ViT~\cite{clamp-vit}, which do not quantize non-linear layers, IPTQ-ViT achieves comparable accuracy. For instance, on Swin-T (W6A6), IPTQ-ViT is only -0.4 box AP and -0.3 mask AP lower than RepQ-ViT~\cite{repq-vit}. As discussed in Section~\ref{sec:motivation}, we observed severe accuracy degradation also on object detection when directly applying QAT-based approximation functions to PTQ. I-ViT$^\ast$ and I-BERT$^\ast$ show average drops of 42.55 box AP and 36.63 mask AP under W4A8 compared to W8A8.

\begin{table}[h]
  \centering
  \resizebox{1\columnwidth}{!}{%
  \begin{tabular}{ccccccccc}
    \toprule
    \multicolumn{2}{c}{\textbf{Cascade Mask R-CNN}} & \multicolumn{2}{c}{\textbf{W8A8}} & \multicolumn{2}{c}{\textbf{W6A6}} & \multicolumn{2}{c}{\textbf{W4A8}} \\
    \cmidrule(lr){3-4} \cmidrule(lr){5-6} \cmidrule(lr){7-8} 
    \textbf{Model} & \textbf{Method} &
    \textbf{AP\textsuperscript{box}} & \textbf{AP\textsuperscript{mask}} & 
    \textbf{AP\textsuperscript{box}} & \textbf{AP\textsuperscript{mask}} & 
    \textbf{AP\textsuperscript{box}} & \textbf{AP\textsuperscript{mask}}
    \\
    \midrule
    \multirow{9}{*}{Swin-S} & Baseline (FP)                  & 51.8 & 44.7 & 51.8 & 44.7 & 51.8 & 44.7 \\
    \cmidrule(lr){2-8}
                            & FQ-ViT~\cite{fq-vit}           & 50.8 & 44.1 & N/A  & N/A  & 48.2 & 41.3 \\
                            & I-ViT$^\ast$                   & 49.1 & 42.4 & 1.2  & 1.0  & 0.3  & 0.3 \\
                            & I-BERT$^\ast$                  & 50.4 & 43.5 & 41.7 & 35   & 14.9 & 13.6 \\
                            & PSAQ-ViT V2~\cite{psaq-vit-v2} & 50.9 & 44.1 & N/A  & N/A  & 47.9 & 41.4 \\
                            & PTQ4-ViT~\cite{repq-vit}       & 20.8 & 18.7 & 12.5 & 10.8 & 38.5 & 33.8 \\
                            & RepQ-ViT~\cite{repq-vit}       & 51.6 & 44.6 & \textbf{51.4} & \textbf{44.6} & \textbf{49.2} & \textbf{42.8} \\
                            & CLAMP-ViT~\cite{clamp-vit}     & 51.4 & 44.6 & N/A  & N/A  & 48.5 & 42.2 \\
                            & \textbf{IPTQ-ViT}       & \textbf{51.8} & \textbf{44.8} & \textbf{51.4} & \textbf{44.5} & 48.2 & 41.8 \\
    \midrule
    \multirow{7}{*}{Swin-T} & Baseline (FP)                  & 50.4 & 43.7 & 50.4 & 43.7 & 50.4 & 43.7 \\
    \cmidrule(lr){2-8}
                            & FQ-ViT~\cite{fq-vit}           & 49.7 & 43.3 & N/A  & N/A  & N/A  & N/A \\
                            & I-ViT$^\ast$                   & 48.3 & 42.7 & 2.9  & 2.6  & 1.3  & 1.2  \\
                            & I-BERT$^\ast$                  & 49.7 & 43.1 & 40.9 & 35.5 & 10.8 & 10.1 \\
                            & PTQ4-ViT~\cite{repq-vit}       & 40.3 & 35.6 & 14.7 & 13.6 & 25.3 & 22.7 \\
                            & RepQ-ViT~\cite{repq-vit}       & NA   & NA   & \textbf{50.0} & \textbf{43.5} & NA   & NA   \\    
                            & \textbf{IPTQ-ViT}      & \textbf{50.4} & \textbf{43.7} & 49.6 & 43.2 & \textbf{43.1} & \textbf{37.7} \\
    \bottomrule
  \end{tabular}
  }
  \caption{Object detection on COCO. ${}^{\ast}$ indicates results reproduced using official code. \textbf{Bold} indicates the top performance methods.}
  \label{tab:swin_results_od}
\end{table}

\subsection{Quantization Runtime}

\begin{figure}[h]
\centering
\includegraphics[scale=0.31]{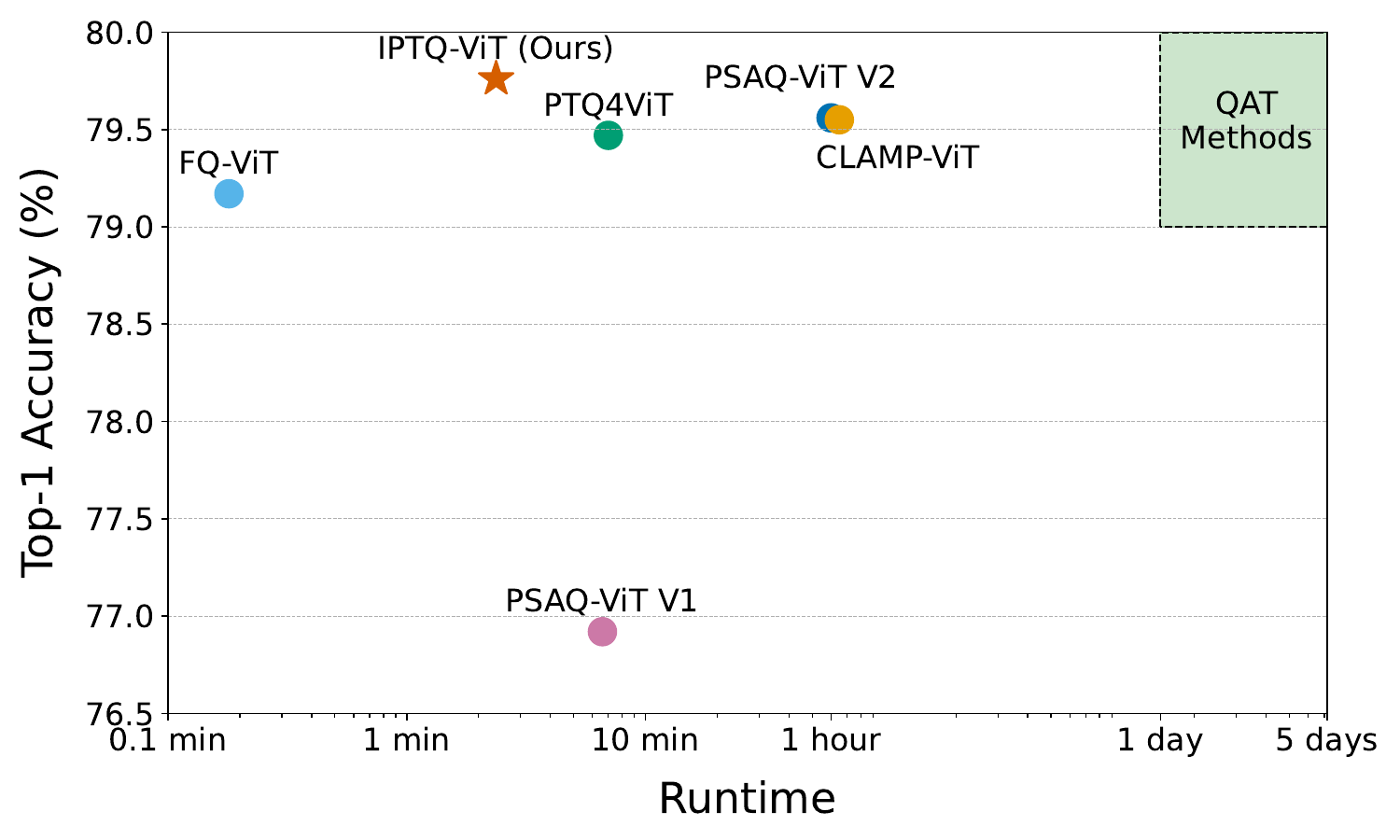}
\caption{Quantization runtime on \emph{DeiT-S} under W8A8, measured on a single NVIDIA RTX 3090 GPU. Times exclude inference.}
\label{fig:runtime}
\end{figure}

In Fig.~\ref{fig:runtime}, we report the quantization runtime and ImageNet-1K top-1 accuracy of IPTQ-ViT compared with prior PTQ methods on DeiT-S (W8A8). The runtime measures only the quantization process, excluding inference. IPTQ-ViT completes quantization in 2.37 minutes and achieves higher accuracy than all baselines. In contrast, PSAQ-ViT V1~\cite{psaq-vit-v1} and CLAMP-ViT~\cite{clamp-vit} require extra procedures such as synthetic data generation, analysis and PTQ calibration, which increase overhead. FQ-ViT~\cite{fq-vit} records the shortest runtime, its accuracy is noticeably lower than IPTQ-ViT. The results of additional models are shown in Appendix~\textcolor{wacvblue}{E}.

\news{
    \subsection{Latency Evaluation}
    \label{sec:experiments_latency}
    Tab.~\ref{tab:latency} reports the latency of DeiT-T, DeiT-S, and DeiT-B under W8A8. For comparison, we also include the original I-ViT~\cite{i-vit} results on an RTX~2080Ti GPU and our re-implementation I-ViT$^\S$ on the same RTX~3090 GPU. IPTQ-ViT achieves a comparable latency to I-ViT$^\S$ on a 3090 GPU, while maintaining +0.10\%p higher top-1 accuracy on DeiT-B without retraining. It consistently delivers a 1.9–2.6$\times$ speedup over the FP32 baseline, with DeiT-B reduced from 18.7\,ms to 7.11\,ms (2.63$\times$ faster). These results demonstrate that the \textit{Unified Metric}-based assignment in the IPTQ-ViT pipeline not only reduces theoretical latency but also translates into real efficiency gains on actual hardware.
}

\begin{table}[h]
\centering
\resizebox{\columnwidth}{!}{%
\begin{tabular}{llccc}
\toprule
\textbf{Method} & \textbf{GPU} & \textbf{DeiT-T (ms)} & \textbf{DeiT-S (ms)} & \textbf{DeiT-B (ms)} \\
\midrule
Baseline (FP)       & 3090   & 3.98 & 6.58 & 18.7 \\
I-ViT~\cite{i-vit}  & 2080Ti & 1.61 & 2.97 & 7.93 \\
I-ViT$^\S$      & 3090   & 1.72 (\textcolor{ForestGreen}{\textbf{$\times$2.31}}) & 2.97 (\textcolor{ForestGreen}{\textbf{$\times$2.21}}) & 6.99 (\textcolor{ForestGreen}{\textbf{$\times$2.68}}) \\
\midrule
Ours                & 3090   & 1.73 (\textcolor{ForestGreen}{\textbf{$\times$2.30}}) & 3.46 (\textcolor{ForestGreen}{\textbf{$\times$1.90}}) & 7.11 (\textcolor{ForestGreen}{\textbf{$\times$2.63}}) \\
\bottomrule
\end{tabular}%
}
\caption{
\news{End-to-end latency (ms) with batch size 8. All results are on an RTX 3090 GPU except the original I-ViT~\cite{i-vit} (2080 Ti).}}
\label{tab:latency}
\end{table}

\section{Ablation Study}
\label{sec:ablation_preliminary}
We evaluate the individual and combined effects of our three proposed methods: (1) \textit{Data-aware Poly-GELU}, (2) \textit{Efficient Bit-Softmax}, and (3) \textit{Unified Metric}. All experiments follow the IPTQ-ViT pipeline (Figure~\ref{fig:overal_pipeline}) and analyze accuracy changes across different ViTs. Results are summarized in Table~\ref{tab:metric_ablation_summary}.
First, replacing SQNR with \textit{Unified Metric} (“Legacy/Unified”) consistently improves accuracy over “Legacy/SQNR”, confirming its effectiveness in guiding approximation selection. "Legacy" denotes only using existing approximation functions.
Next, expanding the search space with our proposed functions (“Extended \{Softmax or GELU\}/SQNR”) further boosts accuracy beyond the “Legacy” setting and even outperforms FQ-ViT~\cite{fq-vit} on most models. Notably, adding only \textit{Efficient Bit-Softmax} yields a +0.47\%p gain on DeiT-B over FQ-ViT, showing its standalone benefit.
When both Poly-GELU and Bit-Softmax are included (“Extended/SQNR”), performance improves over using either alone, demonstrating their complementary nature. Finally, combining the full search space with \textit{Unified Metric} (“Extended/Unified”) delivers the best results across all models.

\section{Acknowledgement}
This work was supported by Korea Research Institute for defense Technology planning and advancement (KRIT) grant funded by the Korea government (DAPA (Defense Acquisition Program Administration)) (No.KRIT-CT-22-040, Heterogeneous Satellite constellation based ISR Research Center, 2025). \\
This work was supported by Institute of Information \& communications
Technology Planning \& Evaluation (IITP) grant funded by the Korea
government (MSIT) (No.RS-2023-00277060, Development of open edge AI SoC hardware and software platform)

\section{Conclusion}
\label{sec:conclusion}

We presented IPTQ-ViT, the first PTQ framework that enables fully integer-only vision transformers, a goal previously infeasible with conventional PTQ. 
By introducing tailored approximations for non-linear layers and a unified metric for effective function assignment, IPTQ-ViT delivers accurate and efficient integer-only inference without retraining. 
Experiments demonstrate that it outperforms prior PTQ baselines, achieves accuracy on par with integer-only QAT methods, and offers strong practicality for deployment with demonstrated latency.





{
    \small
    \bibliographystyle{ieeenat_fullname}
    \bibliography{main}
}

\end{document}